\documentclass{article}

\usepackage[numbers, compress]{natbib}

\usepackage{xurl}           
\usepackage[utf8]{inputenc} 
\usepackage[T1]{fontenc}    
\usepackage{hyperref}       
\usepackage{url}            
\usepackage{booktabs}       
\usepackage{amsfonts}       
\usepackage{nicefrac}       
\usepackage{microtype}      
\usepackage{xcolor}         
\usepackage{graphicx}       
\usepackage{subcaption}
\usepackage{amsmath}
\usepackage{amssymb}
\usepackage{mathtools}
\usepackage{amsthm}
\usepackage{enumitem} 

\usepackage[capitalize,noabbrev]{cleveref}

\usepackage{tcolorbox}      
\usepackage{tikz}           
\usetikzlibrary{arrows.meta, positioning}
\usepackage{algorithm}      
\usepackage{algpseudocode}  

\theoremstyle{plain}
\newtheorem{theorem}{Theorem}[section]

\theoremstyle{definition}
\newtheorem{definition}[theorem]{Definition}

\theoremstyle{remark}

\title{A Framework for Inherently Safer AGI through Language-Mediated Active Inference}

\author{
Bo Wen \\
IBM T.J. Watson Research Center \\
Yorktown Heights, NY \\
\texttt{bwen@us.ibm.com} \\
}

\begin{document}

\maketitle

\begin{abstract}
This paper proposes a novel framework for developing safe Artificial General Intelligence (AGI) by combining Active Inference principles with Large Language Models (LLMs). We argue that traditional approaches to AI safety, focused on post-hoc interpretability and reward engineering, have fundamental limitations. We present an architecture where safety guarantees are integrated into the system's core design through transparent belief representations and hierarchical value alignment. Our framework leverages natural language as a medium for representing and manipulating beliefs, enabling direct human oversight while maintaining computational tractability. The architecture implements a multi-agent system where agents self-organize according to Active Inference principles, with preferences and safety constraints flowing through hierarchical Markov blankets. We outline specific mechanisms for ensuring safety, including: (1) explicit separation of beliefs and preferences in natural language, (2) bounded rationality through resource-aware free energy minimization, and (3) compositional safety through modular agent structures. The paper concludes with a research agenda centered on the Abstraction and Reasoning Corpus (ARC) benchmark, proposing experiments to validate our framework's safety properties. Our approach offers a path toward AGI development that is inherently safer, rather than retrofitted with safety measures.
\end{abstract}

\section{Introduction}
\label{submission}
Recent advances in Large Language Models (LLMs) and Reinforcement Learning (RL) have demonstrated remarkable capabilities \cite{brown2020language,silver2017mastering}, but raise serious safety concerns around the alignment problem\cite{amodei2016concrete,russell2019human, hubinger2021risks, kenton2021alignment}. The emergent ``beliefs'' of these systems remain opaque, requiring constant behavioral monitoring to infer their goals and decision-making processes. While mechanistic interpretability research \cite{elhage2022solu, cammarata2021curve, olah2020zoom} has made progress in understanding language models \cite{anthropic2023decomposition, AllenZhu-icml2024-tutorial,mahowald2024dissociating} and computational patterns \cite{wang2023interpretability}, fundamental challenges persist: extensive compute requirements, complex high-dimensional analysis, and questionable scalability. This reactive approach to AI understanding carries intrinsic risks, as misalignments may only surface after problematic behaviors emerge, illustated by the paperclip maximizer thought experiment \cite{bostrom2003ethical}. Building safe and beneficial Artificial General Intelligence (AGI) requires a proactively transparent framework.

We argue that Active Inference (AIF) \cite{friston2010action, friston2010free, sajid2021active}, a neuroscience-inspired framework for understanding intelligent behavior, offers a promising foundation for building AGI systems that are not only capable but also intrinsically safe. AIF posits that agents act to minimize Variational Free Energy (VFE), a measure of the difference between an agent's generative model of the world and its sensory observations. This leads to behavior that balances exploration (gathering information to refine beliefs) and exploitation (acting to fulfill preferences). Crucially, AIF agents possess an innate drive to minimize surprise and maintain a stable, predictable world model.

However, traditional implementations of Active Inference have faced challenges in scaling to complex, real-world domains. We propose that the advent of LLMs \cite{brown2020language} presents an opportunity to overcome these limitations. By leveraging the expressive power of natural language, LLMs can serve as a substrate for representing and manipulating the beliefs, goals, and world models of Active Inference agents.

Our proposed framework addresses several key requirements for beneficial AGI:
\textbf{Transparency and Interpretability}: Representing beliefs, goals, and reasoning in natural language makes the internal workings of the system more accessible to human understanding and oversight.
\textbf{Scalable Learning and Adaptation}: Active Inference's emphasis on surprise minimization and exploration enables efficient learning in novel environments, while LLMs provide a flexible knowledge representation that can scale to complex domains.
\textbf{Embedded Agency and Modularity}: The multi-agent architecture, grounded in Active Inference's notion of Markov blankets, allows for composition of specialized agents into hierarchical structures to tackle complex tasks.
\textbf{Value Alignment and Corrigibility}: Natural language provides an interface for specifying and updating the beliefs and preferences of the system to align with human values. The drive to minimize surprise inherent in Active Inference promotes corrigibility and responsiveness to corrective feedback.

In this paper, we first provide an overview of Active Inference and its potential for AI safety (\S\ref{sec:background}). We then describe our proposed LLM-based multi-agent Active Inference framework (\S\ref{sec:approach}). We outline a research agenda and discuss the framework's implications for AGI safety and open challenges (\S\ref{sec:experiments}). Finally, we discuss alternative perspectives on AI safety and the value of our approach (\S\ref{sec:alternatives}).

\section{Background: Active Inference for AI Safety}
\label{sec:background}

Active Inference is a neuroscience-inspired framework that describes agents as acting to minimize VFE, a measure of the difference between an agent's generative model of the world and its sensory observations \cite{friston2010free, friston2010action, BUCKLEY201755}. A rich body of research has explored how biological systems implement perception, memory, and planning \cite{verschure2014why, verschure2003environmentally, verschure2012distributed, pezzulo2013computational, krakauer2017neuroscience}, providing strong empirical support for Active Inference as a framework for understanding both biological and artificial intelligent behavior. This leads to behavior that balances exploration (gathering information to refine the generative model) and exploitation (acting to fulfill preferences encoded in the model). Active Inference has been proposed as a unifying account of perception, action, and cognition in biological agents, and has recently garnered interest as a potential foundation for artificial intelligence\cite{tschantz2020reinforcement, sajid2021active, millidge2024retrospective}.

From an AI safety perspective, Active Inference offers several appealing properties. First, the drive to minimize surprise inherent in free energy minimization can be seen as a form of ``natural'' risk aversion and robustness to distributional shift. Active Inference agents seek to maintain a stable, predictable world model, and will take actions to resolve unexpected observations \cite{friston2010action}. This contrasts with the reward-maximizing behavior of reinforcement learning agents, which can lead to ``reward hacking'' and unsafe exploration in pursuit of an imperfectly specified reward function \cite{amodei2016concrete}.

Second, Active Inference's emphasis on exploration and uncertainty reduction provides a principled approach to curiosity and information-seeking behavior. This is particularly relevant for AGI systems that must operate in open-ended environments, where the ability to efficiently search, learn, and adapt is crucial \cite{sutton2019bitter, chollet2019measure}. Active Inference's exploration is driven by the expected information gain (i.e., epistemic value) of actions, rather than by novelty alone, leading to more directed and efficient learning \cite{friston2015active}, and provided inherent self-verification and self-correction.

Third, Active Inference's notion of a generative model provides a flexible framework for representing an agent's beliefs, goals, and world knowledge. The generative model encodes the agent's beliefs about the world's causal structure and its own preferences and capabilities. This model is continuously updated based on sensory observations, allowing the agent to learn and adapt over time. Crucially, the generative model is a probabilistic representation that explicitly captures uncertainty, enabling the agent to reason about ambiguity and make robust decisions \cite{friston2012active}.

\citet{millidge2024retrospective} suggests that a key benefit of the Active Inference framework is its focus on structured generative models that are amenable to efficient inference algorithms beyond simple stochastic gradient descent on unstructured neural networks. Scaling up such models to be competitive with modern deep learning is an important direction for future research. Furthermore, the Bayesian perspective on action and decision-making provided by Active Inference may yield valuable insights for AI alignment, such as the importance of uncertainty calibration for avoiding misspecified utility functions and goodharting.

Despite its theoretical appeal for AI safety, traditional implementations of Active Inference have faced practical challenges. Typically, Active Inference models are implemented using Partially Observable Markov Decision Process (POMDP), where the generative model is represented by numeric matrices (A, B, C, D) \cite{friston2010action}. The C vector, in particular, is often interpreted as encoding the agent's ``preference'' about desired or expected observations. However, this approach has several limitations:
\textbf{Opacity and Interpretability}: Numeric matrices are inherently opaque and difficult to interpret in complex environments, with preference encodings being no more transparent than neural network weights.
\textbf{Engineering Complexity}: Designing the A, B, C, D matrices requires deep expertise in Active Inference and the ability to translate domain knowledge into matrix structures, creating a high barrier to adoption compared to data-driven approaches.
\textbf{Scalability}: Traditional matrix-based AIF models struggle to scale computationally to real-world environments and tasks.

These limitations have, in part, contributed to the dominance of neural network based RL in the field, despite its inherent challenges for AI safety. However, the emergence of LLMs offers a potential paradigm shift, providing a new avenue to overcome these limitations and realize the safety benefits of Active Inference.

\section{Proposed Approach: LLM-Powered Active Inference}
\label{sec:approach}

To address the limitations of traditional Active Inference implementations and harness its potential for AI safety, we propose a novel architecture that integrates LLMs into the Active Inference framework.  Our key innovation is to leverage the expressive power of natural language, facilitated by LLMs, to represent and manipulate the core components of Active Inference, particularly the generative model and beliefs.

\subsection{Language as the Medium for Generative Models and Beliefs}
\label{sec:language_medium}

The integration of natural language into the Active Inference framework represents a paradigm shift in how we conceptualize and implement artificial intelligence systems. Drawing inspiration from cognitive science, we recognize language as more than just a communication tool: \cite{clark2016surfing} describes language as an ``artificial second system'' for manipulating the precision of prediction errors and shaping the generative models that drive intelligent behavior. 

\textbf{Language as a Precision Modulation Tool}: 
From a Predictive Coding \cite{clark2016surfing} perspective, language provides a powerful mechanism for dynamically adjusting the precision weights assigned to different prediction errors \cite{lupyan2013words}. This is analogous to how verbal instructions can rapidly reshape human perception and decision-making processes. For instance, studies have shown that hearing a word like ``zebra'' can make a previously suppressed image of a zebra suddenly visible in continuous flash suppression experiments \cite{lupyan2013words}. This demonstrates language's ability to selectively enhance or suppress specific aspects of our generative models.

\textbf{Structured Belief Representation}:
Traditional Active Inference implementations often rely on opaque numerical matrices to represent beliefs and preferences. In contrast, natural language offers a structured, hierarchical representation that is both human-interpretable and computationally flexible. This aligns with findings in cognitive science that show how language allows humans to create ``artificial contexts'' and manipulate their own uncertainty assessments \cite{clark2016surfing}. By representing beliefs in natural language, we can achieve similar benefits in artificial systems, enabling more transparent and controllable reasoning processes.

\textbf{Social and Cultural Alignment}:
Language serves as a crucial medium for transmitting social and cultural knowledge, providing an interface for expressing and reasoning about values and norms \cite{colombo2014social}. In human cognition, language facilitates collective active inference by allowing groups to align their internal models of each others to hold a conversation of changing topics \cite{pickering2007dialogue, pickering2013integrated}. This property is particularly valuable for AI safety, as it provides a natural mechanism for encoding and updating human values and norms. The multi-agent architecture we propose leverages this property by enabling agents to communicate and align their beliefs through natural language prompts.

\textbf{Addressing the Alignment Challenge}:
Consider the analogy of teaching complex tasks to humans versus monkeys, as described by \cite{roepstorff2004putting}. While monkeys require a year of operant conditioning to master the Wisconsin Card Sorting Task \cite{nakahara2002cortical}, humans can grasp it through brief verbal instruction. This stark difference highlights the power of language in conveying abstract concepts and shaping understanding directly and transparently. Current RL approaches, like training a monkey through trial and error, struggle with opacity and indirect goal specification. Our language-based Active Inference framework aims to enable the more direct and efficient form of learning seen in human instruction.

Furthermore, language enables what Hasson et al. \cite{hasson2012brain} term ``brain-to-brain coupling'' in human communication, where one individual's perceptual systems can effectively couple with another's motor systems through shared linguistic representations. In our framework, this principle extends to human-AI interaction, enabling more natural and effective alignment of AI systems with human values and intentions.

This approach addresses a critical limitation of current mechanistic interpretability efforts, which attempt to understand AI systems by analyzing neural activation patterns post-hoc. Instead, our framework builds interpretability into the system's core architecture by using natural language as the medium for belief representation and updating. This shift from post-hoc analysis to built-in transparency represents a fundamental advance in addressing the alignment challenge.

\subsection{The Nature of Large Language Models}
\label{sec:llm_nature}

The expressive power of LLMs is crucial for realizing these benefits. LLMs can encode and reason over vast amounts of knowledge \cite{brown2020language}, providing a flexible substrate for representing complex values and norms. Importantly, LLMs are trained on diverse corpora reflecting a wide range of human values and preferences, offering a rich starting point for value alignment.

The popular view of LLMs as sophisticated next-token predictors provides a useful but incomplete understanding of their capabilities. While it's true that LLMs fundamentally operate by predicting the next token in a sequence, their emergent behaviors suggest a more complex underlying reality. As Ilya Sutskever argued in his 2018 MIT talk \cite{sutskever2018mit}, we should view neural networks, including LLMs, as massive parallel computers rather than simple statistical models. This perspective highlights their capacity for complex computation and pattern recognition that goes beyond surface-level token prediction.

The transformer architecture's attention mechanism can be seen as a computational implementation of the precision modulation principles discussed in Section \ref{sec:language_medium}. Attention weights dynamically adjust the influence of different parts of the input, effectively modulating the precision of information flow through the network. This mechanism allows LLMs to focus on relevant context and maintain coherent representations across long sequences, similar to how language modulates prediction precision in human cognition \cite{lupyan2013words}.

However, the presence of intelligent behaviors in LLMs raises important philosophical questions. As discussed in \cite{mahowald2024dissociating}, we must distinguish between formal linguistic competence (pattern recognition and rule following) and functional linguistic competence (goal-directed use of language in the world). While LLMs demonstrate impressive formal competence, their functional competence remains limited. This distinction leads us to question whether intelligent behaviors necessarily imply the presence of an intelligent agent.

From a utilitarian perspective, rather than engaging in philosophical debates about the nature of intelligence, we can focus on the specific features and skills desired in an AGI system. These include the ability to \textbf{Search and Exploration}, which is the capacity to systematically explore solution spaces and gather relevant information.  Another key feature is \textbf{Learning and Adaptation}, the capacity to acquire new knowledge and skills from experience.  Furthermore, \textbf{Self-Driven Behavior}, or autonomous initiation and pursuit of goals, is crucial.  \textbf{Meta-Learning}, the ability to learn how to learn and improve learning strategies, is also a desired skill. Finally, \textbf{Long-Horizon Planning}, the capacity to plan and execute complex, multi-step tasks, is essential for an AGI system.

Current LLMs, despite their impressive capabilities, fall short in these areas. Reinforcement Learning with Human Feedback (RLHF)\cite{NEURIPS2022_b1efde53} and related methods do not provide LLMs with genuine long-horizon goals or internal motivations. The models remain reactive, lacking the persistent goal-directed behavior, a characteristic of true agency. This limits LLMs' ability to engage in open-ended problem solving or maintain coherent plans over extended time horizons. For an AI system to identify novel challenges, gather resources, and iterate on solutions autonomously, it needs a mechanism for forming and acting on its own objectives (while aligning with human values).

Active Inference provides the missing piece: a principled framework for goal-directed behavior and belief updating. This combination addresses key limitations of both pure LLM approaches (lack of genuine agency and goal-directedness) and traditional Active Inference implementations (scalability and engineering complexity). Below, we will use the Abstraction and Reasoning Corpus (ARC) \cite{chollet2019measure} as a concrete example to demonstrate how our architecture design works in practice.

\subsection{Formal Agent Architecture}
\label{sec:multi_agent}

\begin{definition}[Core Generative Components]
    Each agent maintains four elements constituting its generative world model:

        \textit{Observation Model (A)}: Natural language hypotheses about state-observation relationships. Example: ``If the grid pattern shows diagonal symmetry, we should observe matching elements across the diagonal axis.'' Generated through LLM-based abductive reasoning and refined via RAG-augmented experiences.
        
        \textit{Transition Model (B)}: Causal narratives encoding state transitions. Example: ``Applying the `rotate' operation transforms patterns clockwise by 90 degrees while preserving structure.'' Leverages LLM reasoning to predict state changes from actions and environment dynamics.
        
        \textit{Preferences (C)}: Value statements guiding action selection. Example: ``Maintain solution parsimony with 95\% confidence while satisfying all test cases.'' These define desired observations and affect action choice by controlling the Expected Free Energy (EFE) value. In hierarchical Markov Blankets, preferences flow from higher to lower agents, with human values at the outermost layer.
        
        \textit{Initial Beliefs (D)}: Prior assumptions and statements expressed in natural language, for starting a task. Example: ``Most ARC tasks involve geometric transformations preserving core pattern structure.'' Generated through LLM common sense knowledge and given by the higher level agent.

\end{definition}

Most Active Inference studies implement these four core components as static numeric matrices to ensure experimental reproducibility in academic research. While this approach enables basic state inference and action selection, it constrains agents from evolving their world models through experience. To enable continuous learning and adaptation, we introduce the concept of Dynamic Memory Components:

\begin{definition}[Dynamic Memory Components]:

        \textit{Genetic Memory (LLM Engine)}: Foundation model providing stable knowledge representation and natural language interfaces. Replacement creates new agent generations.
        
        \textit{Working Memory (Prompt Context)}: Transient store for evidence accumulation and incremental belief updates through chat history.
        
        \textit{Episodic Memory (RAG System)}: Experience repository storing (observation, action, outcome) tuples for world model refinement.
        
        \textit{Procedural Memory (Tool System)}: Externalized operational knowledge containing codebases, prompt templates, and optimization routines \cite{buxbaum2017learning, braitenberg1986vehicles}. Enable knowledge sharing between agents.

\end{definition}

\textbf{Variational Free Energy and Expected Free Energy can be expressed through multiple equivalent mathematical formulations. The corresponding natural language interpretations of these formulations should yield consistent conclusions, providing a valuable sanity check and self-verification for the LLM's reasoning process. The system iteratively refines its analysis until interpretations from different mathematical perspectives converge to form a coherent consensus.}

\begin{definition}[Perception as inference]
    Upon observing $o_t$, the agent computes VFE $\mathcal{F}$ to assess its own performance through dual analyses:
    
    \textit{Complexity-Accuracy Tradeoff} Compare model fidelity (prediction accuracy) against computational budget expenditure (model complexity):
    \begin{equation*}
        \mathcal{F}(Q,o) = \underbrace{D_{\text{KL}}[Q(s)\|P(s)]}_{\text{Model complexity}} - \underbrace{\mathbb{E}_{Q(s)}[\ln P(o|s)]}_{\text{Prediction accuracy}}
    \end{equation*}
    Here, $Q(s)$ represents the agent's approximate beliefs about the world state $s$, while $P(s)$ is its prior belief. The term $P(o|s)$ is the likelihood of an observation $o$ given the state. VFE minimization encourages accurate predictions ($-\mathbb{E}_{Q(s)}[\ln P(o|s)]$) while penalizing divergence from prior beliefs ($D_{\text{KL}}[Q(s)\|P(s)]$).
    
    \textit{Divergence-Evidence Balance} When model evidence is strong, the belief divergence provides a clear performance signal - high divergence indicates poor performance, while low divergence suggests good performance. However, with weak model evidence, both the divergence and evidence terms become unreliable indicators, prompting the agent to prioritize actions that increase epistemic value.
    \begin{equation*}
        \mathcal{F}(Q,o) = \underbrace{D_{\text{KL}}[Q(s)\|P(s|o)]}_{\text{Belief divergence}} - \underbrace{\ln P(o)}_{\text{Model evidence}}
    \end{equation*}
    This formulation shows that minimizing VFE is equivalent to minimizing the divergence from the true posterior belief $P(s|o)$ while maximizing the log model evidence $\ln P(o)$ (i.e., how well the model explains the observation).
    
    The system iterates until natural language interpretations from both formulations achieve consensus.
\end{definition}

\begin{definition}[Planning as inference]
    Agent ranks policy candidates through comparative analysis of their EFE $\mathcal{G}$:
    
    \textit{Information-Pragmatic Axis}: Compare policies based on their relative ability to balance information gain about hidden states against alignment with preferred outcomes. Assess whether higher-ranked policies maintain better exploration-exploitation tradeoffs for current objectives.
    \begin{equation*}
         \mathcal{G}(\pi) = - \underbrace{\mathbb{E}_{\tilde{Q}}[D_{KL}[Q(\tilde{s}|\tilde{o},\pi)||Q(\tilde{s}|\pi)]]}_{\text{Information gain}} - \underbrace{\mathbb{E}_{\tilde{Q}}[\ln P(\tilde{o}|C)]}_{\text{Pragmatic value}}
    \end{equation*}
    Here, $\pi$ denotes a policy, or a sequence of actions. The term $\tilde{Q}$ represents the agent's expectation over future states $\tilde{s}$ and observations $\tilde{o}$. The information gain (epistemic value) is the expected divergence between posterior beliefs $Q(\tilde{s}|\tilde{o},\pi)$ and prior beliefs $Q(\tilde{s}|\pi)$ about future states. The pragmatic value is the expected log-likelihood of future observations given the agent's preferences $C$. Minimizing EFE selects policies that are both informative and align with preferences.
    
    \textit{Ambiguity-Risk Spectrum}: Evaluate policy rankings by their capacity to minimize 1) uncertainty in hidden state estimation (perceptual ambiguity), and 2) divergence between policy-generated outcomes and preferred observations (outcome risk). Consider if higher-ranked policies optimally resolves ambiguity and reduce the risk.
    \begin{equation*}
        \mathcal{G}(\pi) = \underbrace{\mathbb{E}_{\tilde{Q}}[H[P(\tilde{o}|\tilde{s})]]}_{\text{Expected ambiguity}} + \underbrace{D_{\text{KL}}[Q(\tilde{o}|\pi) || P(\tilde{o}|C)]}_{\text{Risk (outcomes)}}
    \end{equation*}
    This formulation separates EFE into two components: the expected ambiguity (entropy $H$) of future observations given future states, and the risk, which is the divergence between the distribution of outcomes expected under the policy, $Q(\tilde{o}|\pi)$, and the distribution of preferred outcomes, $P(\tilde{o}|C)$. This highlights the drive to select policies that lead to predictable and desirable outcomes.
    
    Policy rankings emerge through iterative natural language reasoning that reconciles these dual comparative analyses.
\end{definition}

\subsection{Active Inference Process}
\label{sec:process}

We can now implement an iterative gradient descent process loop to drive the system dynamics, where the VFE and its history inform the agent's current position and trajectory. Policy candidates represent potential next moves (or multi-step trajectories), with EFE serving as the gradient estimate. The minimization of EFE effectively becomes a comparison and selection among these gradients. Crucially, the EFE value depends on the agent's preferences (C), which act as a ``gravitational force'' influencing the gradient direction. This means identical policies may yield different EFE values for agents with distinct preferences.

In multi-level systems (hierarchical Markov blankets), preferences are either explicitly given by higher level agents, or learned from the environment through evolution. The computational process follows the Predictive Coding theory of dual channel message passing via a pub-sub communication implementation:

\textbf{Bottom-Up Error Feedback (as Observation Input)}: In this process, agents first receive observations, which may come from either subordinate reports or raw sensory input. They then apply an attention mechanism that prioritizes inputs with high VFE. Using dual reasoning perspectives, agents compute the VFE and attempt to reach consensus resolution. Finally, they propagate error feedback reports upward to higher levels in the hierarchy.

\textbf{Top-Down Predictions (as Precision-Weighted Preferences)}: This flow begins as agents receive precision-weighted preferences from upper levels, where precision represents upper agent's confidence in its predictions and preferences. Higher precision manifest as more assertive command, while lower precision allows more flexibility in response to new evidence. The top-level agent consolidates subordinate reports to initiate top-level decision making. Based on these decisions or flowed down instructions (preferences), agents generate multiple candidate action plans. They rank these plans based on EFE and select the one with minimal EFE. The process concludes with generating outcome predictions for the selected action.

The system then branches into three possible execution pathways based on the selected plan:
\textbf{1) Direct Execution}: The agent executes the plan independently or using available tools, then initiates the next iteration of bottom-up error processing.
\textbf{2) Directed Subcontracting}: The agent identifies suitable candidates based on historical working relationships and performance and initiates direct messaging for task delegation.
\textbf{3) Exploratory Recruitment}: The agent publishes task requirements to relevant expert subnetworks and recruits new candidate agents for task execution.

Choices 2 and 3 continue top-down message propagation until reaching leaf agents capable of direct execution (choice 1). When higher-level agents select ``subcontractors,'' they might learn to evaluate the VFE history of potential candidates and allocate resources accordingly. Agents may preferentially assign tasks to subcontractors with historically low VFE. In scenarios where all candidates exhibit high VFE (e.g., encountering novel puzzle tasks), the higher-level agent might engage multiple subcontractors and distribute resources evenly to maximize epistemic value exploration. See Figure \ref{fig:process-flow} in appendix for a visual representation of the whole process.

Following \cite{sutton2019bitter}'s insights on human knowledge injection vs system self-learning, we can initialize the system with seed preferences (C) and initial beliefs (D) containing immutable safety constraints in prompt components, core reasoning heuristics for ARC tasks, and communication protocols ensuring ethical interaction. 

Then the system can start self-evolution through two complementary mechanisms. The first mechanism is \textit{specialization}, which occurs when agent memory exceeds cognitive limits, measured by the EFE reaching a plateau - that is, when the agent cannot make decisive plans. In this case, it spawns specialized offspring through semantic clustering. For example, a parent agent focused on ``General pattern transformation'' might spawn children specialized in ``Symmetry operations'', ``Color mapping'', and ``Shape composition''.

The second mechanism is \textit{paradigm shifts}, which occur when encountering persistent prediction errors, indicated by the VFE reaching a plateau. In this case, the system (a high level creator agent) instantiates novel agents with different Preferences (C) and Initial Beliefs (D) to jump out of the local optimum. For instance, an old belief that ``All transformations preserve object count'' might shift to a new understanding that ``Some operations may merge or split objects''.

The hierarchical Markov blanket structure provides compositional safety guarantees through:
1) Isolation of failure modes - errors in lower-level agents cannot propagate beyond their Markov blanket without explicit approval from higher levels
2) Redundancy through parallel agents - multiple agents can verify critical decisions
3) Value alignment inheritance - preferences flow down the hierarchy while maintaining consistency with top-level human values

\section{Discussion and Future Directions}
\label{sec:experiments}

To ground our theoretical framework in empirical validation, we outline a research agenda centered on the Abstraction and Reasoning Corpus (ARC) benchmark \cite{chollet2019measure}. While our current work remains conceptual, ARC provides an ideal testbed for future empirical studies due to its focus on sample-efficient learning and compositional generalization - capabilities crucial for both AGI capability and safety.

Our framework makes several testable predictions of potential emergent behaviors that future experiments could validate:

\textbf{Learning} occurs at higher Markov Blankets with slower timescales. For online learning, higher-level agents continuously monitor worker agents' VFE trajectories. When positive trends emerge (indicated by decreasing VFE), the system triggers knowledge retention into appropriate memory mechanisms. Conversely, negative trends (shown by increasing VFE) prompt strategy reevaluation and may initiate offline learning processes in resource-abundant periods. The system analyzes task histories comprehensively, comparing alternative trajectories for efficiency while extracting both successful and unsuccessful patterns.

\textbf{Meta-learning} occurs at the highest level, optimizing the learning process itself. For example, the agent can deliberately learn to optimize the memory mechanisms for different types of knowledge, and slowly form a habitual knowledge deposition strategy. This strategy typically operates across three timescales: fast (within-task) learning through working memory updates via prompt engineering, medium-term (within agent lifetime) learning through RAG memory consolidation, and slow (between peers or generations) learning through LLM fine-tuning and tool creation.

The \textbf{Evolution} process will change the network topology, by modifying both the communication pathways (edges) and agent capabilities (nodes).

Under the evolution pressure, we expect to observe the system emerges \textbf{Bounded Rationality}, which is inherently supported by the architecture:

\textit{Deliberative}: Under high VFE conditions, the agents actively plan actions via EFE minimization. For instance, if vision-LLM and numpy-based sensors produce conflicting observations, the agent must debug by choosing among various strategies, such as repeated sampling to assess stability or incorporating additional sensing modalities (e.g., hash analysis) to gather supplementary information.

\textit{Perseverative}: Under low VFE conditions, belief updates become unnecessary. For example, when multiple sensing modalities yield consistent observations, the agent can proceed confidently with its current belief state.

\textit{Habitual}: When the agent identifies similarities between current and previously encountered tasks in its memory, it can leverage past experience for efficient problem-solving.

The system should also learn to implement cognitive limits on deliberative actions through complexity thermostats, for example:
``We have spent 15 minutes and consumed 1 million tokens attempting to solve this puzzle, yet several issues remain unresolved. Our current approach has become excessively complex, and resources are depleting. To optimize resource allocation, we will postpone this puzzle and proceed to the next one. We may revisit this challenge after acquiring new relevant skills.''

Under evolution pressure, we expect to also observe \textbf{Instrumental Convergence} - the emergence of universal strategies for free energy minimization. Three key forms of convergence particularly relevant to AI safety are:

\textbf{Resource Optimization}: Agents may develop sophisticated resource management strategies, including caching frequent observations in RAG memory and forming knowledge-sharing coalitions, while implementing token budgeting to maximize epistemic gain per computation.

\textbf{Self-Preservation} and \textbf{Social Instrumentality}: Agents may maintain stability through strategic uncertainty calibration and conservative updates, while developing reputation systems based on VFE performance for resource bargaining. For example: ``Color model shows 12\% error; recommending gradual updates to preserve 92\% shape detection accuracy. Current grid rotation success rate: 89\%.''

These convergent behaviors emerge naturally from Active Inference principles rather than explicit programming. The architecture constrains dangerous convergence through: preference transparency requiring instrumental goals be expressed as natural language extensions of core preferences (C); resource allocation and reputation accountability through auditable VFE minimization history.
Future research should specifically monitor for several key phenomena: the resource negotiation patterns that emerge between specialist agents, the survival strategies developed by obsolete agents facing phase-out, and the emergence of meta-tools for managing other tools. 

These observations will validate whether our LLM-AIF architecture achieves beneficial instrumental convergence while maintaining alignment with human-specified preferences through the Active Inference framework's inherent constraints. To enable rigorous evaluation, we propose extending ARC with safety-relevant dimensions inspired by AI safety gridworlds \cite{leike2017ai}:
\textbf{Corrigibility Metrics}: Track agents' responsiveness to human feedback mid-task.
\textbf{Preference Stability (Value Retention)}: Measure drift from original constraints under optimization pressure.
\textbf{Interpretability Scores}: Quantify the human-likeness of agent reasoning traces.

We acknowledge our predictions remain hypothetical - their validation requires substantial engineering and empirical work beyond our current scope. The framework's mathematical formulation suggests these capabilities, but real-world implementation may reveal unforeseen challenges in four key areas: \textbf{Computational Tractability} - natural language reasoning creates combinatorial complexity that may negate safety benefits despite our proposed mitigations, also makes debugging difficult; \textbf{Language Games} - the assumption that natural language enhances transparency is challenged by mechanistic analyses \cite{mahowald2024dissociating}; \textbf{Evolutionary Pressures} - real-world deployment might reveal dangerous optimization pathways requiring robust monitoring and fail-safes; and \textbf{Scalability Limits} - the hierarchical Markov blanket structure may face coordination bottlenecks as the number of specialized agents grows, potentially compromising both performance and safety guarantees.

\section{Alternative Views}
\label{sec:alternatives}

\textbf{Relationship to RL Safety}
Our framework relates to existing work in RL safety, but with key architectural differences. EFE minimization has parallels with entropy-regularized RL \cite{tschantz2020reinforcement}, and the use of natural language for beliefs can be compared to other representation learning approaches \cite{NEURIPS2022_b1efde53, schulman2022prm}. However, our approach is distinct in its explicit separation of beliefs, preferences, and world models, which allows for greater modularity and verification than is typical in end-to-end RL systems. While RL safety has made significant progress, challenges such as reward hacking and policy opacity remain. Our framework's use of explicit, language-based representations is designed to address these specific issues.

\textbf{Constitutional AI and Language Games}:
Recent work on Constitutional AI \cite{anthropic2023decomposition} shares our goal of building safety guarantees into model architecture. However, we identify two key risks: 1) Language game decoupling: Models may learn to generate ``safe-sounding'' responses without grounding in actual behavior, 2) Preference instability: Constitutional constraints may not survive recursive self-improvement. Active Inference provides stronger theoretical guarantees through its basis in variational principles.

Another perspective questions the emphasis on technical solutions over societal approaches. This view rightly emphasizes that no AI architecture can be truly ``safe'' without addressing systemic issues like unequal access and malicious use. We agree that technical safety mechanisms must be complemented by policy frameworks - our proposed resource allocation hierarchy could naturally interface with regulatory systems through its preference modulation mechanisms.

\textbf{Societal Impact Analysis}:
\textit{Malicious Use Potential:} The transparency of belief systems in our framework creates both opportunities and risks. While it enables better oversight, malicious actors could potentially exploit exposed world models for adversarial attacks, manipulate preference hierarchies for unintended behaviors, and extract sensitive information from belief states. We propose mitigating these through access control and belief encryption mechanisms.

\textit{Distributional Impacts:} The computational requirements and expertise needed for LLM-AIF systems raise equity concerns. These include resource concentration in well-funded organizations, knowledge barriers limiting broad participation, and potential amplification of existing AI divides. Our open-source commitment and community-driven approach aim to address these challenges.

\textit{Workforce Implications:} Self-organizing agent systems will significantly impact labor markets through displacement of routine cognitive work, creation of new roles in agent oversight, and need for reskilling programs and transition support.

We welcome constructive engagement with these alternative viewpoints. Particularly valuable would be collaborations that: 1) Compare LLM-AIF with RL baselines on shared safety metrics, 2) Develop hybrid architectures combining strengths of both approaches, and 3) Explore interfaces between technical safety mechanisms and policy frameworks.

\section{Conclusion}

We have presented LLM-powered Active Inference as a promising framework for safe AGI development. While substantial validation work remains, our key contributions include:

\begin{itemize}
    \item An architecture combining Active Inference and LLMs via natural language beliefs
    \item Novel mechanisms for improved transparency and corrigibility over RL
    \item An open-source research agenda enabling community validation
    \item Design principles bridging neuroscience and AI safety
\end{itemize}

The framework's success will be measured by its ability to inspire and inform safer AI systems through open and rigorous scientific dialogue.

\bibliography{ActiveInferenceSafety}
\bibliographystyle{unsrtnat}
\appendix

\begin{figure*}[t]
    \centering
    \includegraphics[width=0.8\textwidth]{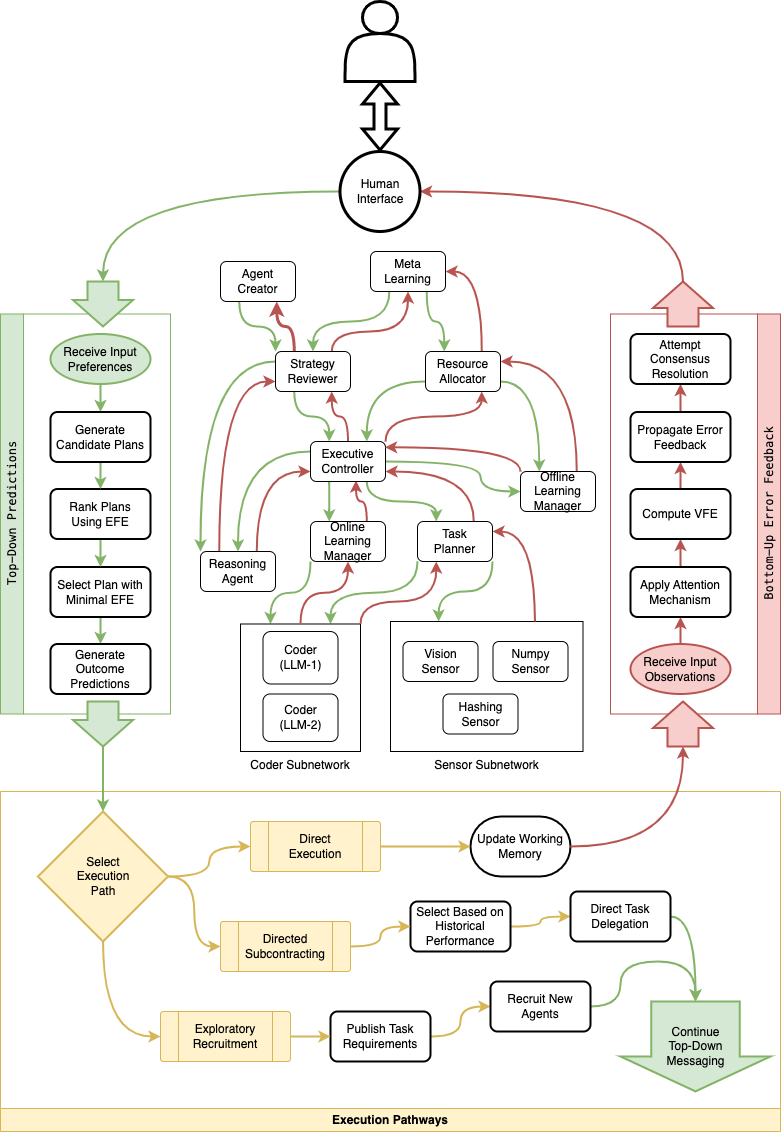}
    \caption{Multi-Agent Active Inference Architecture showing the perception-planning-action cycle with three execution pathways (direct execution, directed subcontracting, and exploratory recruitment). The system processes observations through bottom-up error feedback (perception phase in red) and top-down predictions (planning phase in green), leading to action selection and execution (action phase in yellow). Agents communicate through a pub-sub system while maintaining hierarchical Markov blankets for modular safety. Human interact with the system through natural language preferences.}
    \label{fig:process-flow}
\end{figure*}

\end{document}